\documentclass{article}

\PassOptionsToPackage{numbers, compress}{natbib}
\usepackage[final]{neurips_2024}

\usepackage[utf8]{inputenc} 
\usepackage[T1]{fontenc}    
\usepackage{hyperref}       
\usepackage{url}            
\usepackage{booktabs}       
\usepackage{amsfonts}       
\usepackage{nicefrac}       
\usepackage{microtype}      
\usepackage{xcolor}         

\usepackage{graphicx}
\usepackage{wrapfig} 

\usepackage{graphicx,url}

\usepackage[utf8]{inputenc}

\usepackage{booktabs}
\usepackage{placeins}

\usepackage{graphicx}
\usepackage{caption}
\usepackage{subcaption}

\usepackage{amsmath}
\usepackage{mathtools}

\title{No Argument Left Behind: Overlapping Chunks for Faster Processing of Arbitrarily Long Legal Texts}

\author{Israel Fama\textsuperscript{1}\thanks{Equal contribution.}, Bárbara Bueno\textsuperscript{1*}, Alexandre Alcoforado\textsuperscript{1},
\\
\textbf{Thomas Palmeira Ferraz}\textsuperscript{2}, \textbf{Arnold Moya}\textsuperscript{1}, \textbf{Anna Helena Reali Costa}\textsuperscript{1} 
\\
\\
\textsuperscript{1}Escola Politécnica, Universidade de São Paulo (USP), São Paulo, SP, Brazil
\\
\textsuperscript{2}Télécom Paris, Institut Polytechnique de Paris, Palaiseau, France
\\
\\
\tt{\{israelfama, barbarabueno, anna.reali\}@usp.br}
}

\begin{document} 

\maketitle

\begin{abstract}
  In a context where the Brazilian judiciary system, the largest in the world, faces a crisis due to the slow processing of millions of cases, it becomes imperative to develop efficient methods for analyzing legal texts. We introduce \texttt{uBERT}, a hybrid model that combines Transformer and Recurrent Neural Network architectures to effectively handle long legal texts. Our approach processes the full text regardless of its length while maintaining reasonable computational overhead. Our experiments demonstrate that \texttt{uBERT} achieves superior performance compared to BERT+LSTM when overlapping input is used and is significantly faster than ULMFiT for processing long legal documents. 
\end{abstract}


\section{Introduction}









Legal NLP can be defined as the application of Natural Language Processing (NLP) techniques within the legal domain. This subfield of NLP has been experiencing rapidly growing interest from both academia and industry: \citet{LegalNLP-survey} reports a significant increase in the volume of publications, rising from fewer than 30 papers in 2013 to nearly 120 in 2022. Brazil possesses the largest judiciary system in the world, comprising 18,000 judges distributed across 91 courts. At the time of writing, there are more than 84 million ongoing legal cases \cite{cnj2024}. Despite each judge conclusively adjudicating nearly seven legal cases per day, the average duration of a legal case in Brazil is four and a half years. These numbers indicate both the need and the opportunity for innovative solutions to manage and analyze vast amounts of legal data.

 We turn our focus to Legal Judgment Prediction (LJP), which involves predicting court decisions. Although predicting decisions may be a complex task, we argue it can be reduced to a Text Classification task, which has seen a marked increase in studies ~\cite{survey_on_text_classification}, fueled by advancements in deep learning. In particular, the Transformer architecture emerged as a paradigm shift for many NLP tasks~\cite{paradigm_shifter}. However, it still has limitations when handling long texts, which poses significant challenges in the legal domain, where documents are usually long and complex. 

There is fruitful research being done on enhancing the input size limitation for Transformers, such as Retrieval-Augmented Language Models (RALMs) ~\cite{guu2020realm}. Current retrieval techniques, however, often trust embedding models which also can be sub-optimal when dealing with legal documents, where a single word in the whole document can make a difference. Also, these methods demand substantial computational resources and large document stores to achieve good performance. Other methods combine input in a sequential way, often leveraging properties of Recurrent Neural Networks to process longer sequences ~\cite{long-length_legal_doc_clas}, although those will also usually truncate the text if it is too long. But for documents in the legal domain, such as judicial decisions, most of the documents are usually composed of reasoning from the judge. Therefore, it is of our interest to have a method that uses the full text as input. 

In this paper, we propose \texttt{uBERT}, a hybrid model that combines an encoder-based Transformer with a Recurrent Neural Network, capable of processing long texts. We propose an experimental setup with data from legal decisions, and compare \texttt{uBERT} to baselines BERT+LSTM, Big Bird and ULMFiT in the classification task. Our results show that \texttt{uBERT} slightly outperforms BERT+LSTM as long as overlapping input is introduced. Also, ULMFiT performs better for long texts, but is 4x slower than \texttt{uBERT}.

The remainder of this paper is structured as follows: Sect. \ref{sec:related_works} reviews related work on Legal NLP and long text classification; Sect. \ref{sec:method} outlines our proposal, including the formalization of the target task and the introduction of our model; Sect. \ref{sec:experiments} outlines the experiments we setup to assess our model in terms of performance and efficiency. Finally, we present the results and conclude with a discussion of the findings.


\section{Related Work} \label{sec:related_works}



\subsection{Transformer-Based Approaches for Long Text Processing in Legal NLP}

Enhancing transformer architectures to efficiently process longer texts has become a critical area of research, with significant implications for the legal domain. Longformer ~\cite{beltagy2020longformerlongdocumenttransformer} employs a sparse attention mechanism, extending the input size limit to 4096 tokens, which is eight times the limit of BERT~\cite{BERT}).  
\citet{hoang-etal-2023-viettel} applied this architecture to classify legal texts from the Indian Legal Documents Corpus -- ILDC ~\cite{ildc}, but they did not process the entire text, potentially leading to the loss of relevant information.   

\citet{pappagari2019hierarchicaltransformerslongdocument} introduced Recurrence over BERT -- RoBERT, where longer texts are split into overlapping chunks that are recurrently encoded. Although this approach is conceptually similar to our proposed architecture, a direct comparison is not feasible due to the lack of detailed information on the overlap and recurrence encoding strategies used in RoBERT. Additionally, the authors tested RoBERT on much shorter texts than those in our dataset.

The overlapping algorithm in our approach, \texttt{uBERT}, can be seen as a specific case of the method proposed in SlidingBERT ~\cite{Zhang2023SlidingBertST}, where the sliding window stride is set to $\left \lfloor \text{overlap}/2 \right\rfloor$. Unlike SlidingBERT, which allows tokens to appear in multiple chunks, we limit the overlap so that tokens are covered by at most two chunks. This design choice is motivated not by the language (Portuguese vs. English) but by the goal of reducing computational overhead, ensuring that only adjacent chunks overlap to maintain continuity in context flow.



\citet{brcad-5} introduced BrCAD-5\footnote{ This dataset consists of decisions issued by the Brazilian Federal Small Claims Court (FSCC). These decisions can be appealed to the Appellate Panel (AP), which re-examines the case and either reverses or affirms the initial ruling. Each data point in BrCAD-5 represents the text of a decision issued by the FSCC. The task proposed by the authors is to predict whether the AP will reverse or affirm the initial ruling based on the decision text.}, a dataset designed for Legal Judgment Prediction (LJP), and evaluated three architectures for this task: ULMFiT ~\cite{ULMFiT}, BigBird ~\cite{bigbird}, and BERT+LSTM. ULMFiT, a transfer learning model that fine-tunes a pre-trained language model for downstream NLP tasks, was the only architecture capable of processing the entire text as input. BigBird, a sparse-attention model, addresses the 512-token limit by focusing on subsets of tokens, thereby reducing computational complexity, and was configured to handle texts up to 7,680 tokens. For BERT+LSTM, documents were split into 512-token chunks, with truncation applied to middle chunks if a document required more than 15. 
While similar in approach, \texttt{uBERT} differs from BERT+LSTM in that it uses a chunk overlapping strategy and imposes no limit on the number of chunks, ensuring the entire text is utilized without truncation.

\subsection{Critiques and Limitations in Legal NLP Research}

The legal industry has been slow to adopt NLP advancements, relying heavily on manual work by lawyers. \citet{law_and_nlp} identify a key issue: Legal NLP research often fails to align with the practical needs of legal practitioners. \citet{medvedeva-mcbride-2023-legal} further highlight a significant gap in Legal Judgment Prediction (LJP) research, criticizing the use of poorly designed datasets that rely on biased case facts extracted from judgments. This approach leads to models with overly optimistic performance that offer limited practical value to legal practitioners.

This work aims to bridge the gap between research and practice in the field of Legal NLP. We propose an architecture capable of processing virtually infinite-length legal texts and evaluate it on the BrCAD-5 dataset, which \citet{medvedeva-mcbride-2023-legal} regard as a well-designed benchmark.

\section{Proposal}
\label{sec:method}

Text classification can be formalized as follows.
Given a document $d$ that represents a judicial decision, the goal is to make a prediction $y \in \{0,1\}$, by learning a binary classifier $f$ such that $f(d) = y$. The positive class \( y = 1 \) represents a decision that will be reversed by an Appellate Panel (AP). 
Since legal documents are often long, when using Transformer-based models, conventional approaches usually truncate text from $d$, which can be effective for some semantic tasks \cite{alcoforado2022zeroberto}, but is often sub-optimal for legal text processing tasks ~\cite{pappagari2019hierarchicaltransformerslongdocument}. This can hinder performance on the Legal Judgment Prediction task, since some relevant part of the text may be cut off. 



Believing that the text as a whole is more useful when learning a classifier, we propose  unlimited BERT, or \texttt{uBERT}, an efficient architecture that combines an encoder-based Transformer with a Recurrent Neural Network, utilizing an overlapping algorithm during both training and inference to handle an unlimited number of input tokens. This approach is similar to the BERT+LSTM model used by \citet{brcad-5}, but introduces modifications to maintain local context (through overlapping chunks) and accommodate documents of virtually any size. Although the quadratic memory complexity of the self-attention mechanism presents a challenge for scaling input indefinitely, we leverage the RNN's capacity to process long sequences, enabling it to take chunk embeddings and output a comprehensive document embedding. Several studies, such as done by \citet{hoang-etal-2023-viettel}, have explored the combination of attention mechanisms and recurrence. Our model builds on this concept but applies overlapping during both training and inference, and does not limit the number of chunks processed by the encoder. 

\begin{figure}[h]
    \centering
    \includegraphics[width=0.8\textwidth]{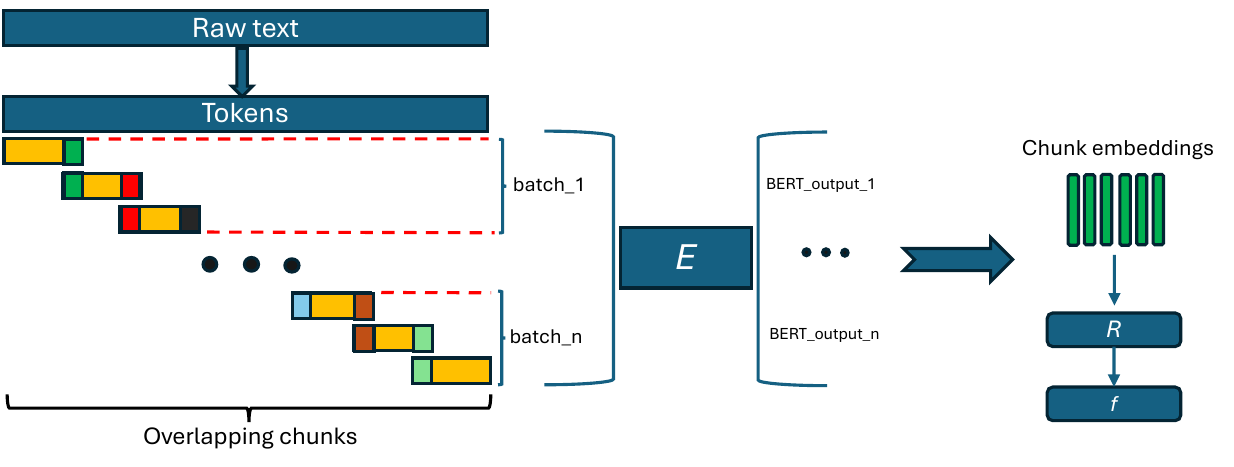}
    \caption{\texttt{uBERT} architecture.}
    \label{fig:arquitetura}
\end{figure}

Figure \ref{fig:arquitetura} depicts the \texttt{uBERT} architecture. It shows key aspects to understand how our model works.

Let $E$ be an encoder-based Transformer, with $dim$ being the dimensionality of the output vector of the final layer, and $R$ be a Recurrent Neural Network. Let $max_{tok}$ be the maximum number of tokens $E$ can process as input. Let $max_{c}$ be the maximum number of chunks of $max_{tok}$ tokens that $E$ can process in parallel with a single run. We split document $d$ into $n$ chunks of size $max_{tok}$ tokens, starting in the first token. For each run, we extract the hidden states from the last four layers of $E$, and concatenate them to form the representation of each chunk. This is based on the idea that different layers capture different linguistic features ~\cite{tenney2019bertrediscoversclassicalnlp}. Specifically, BERT+LSTM, the baseline most similar to our proposed architecture, extracts the hidden states from the last four layers. While other layers could be used for extraction, we retained this approach for consistency in model comparison. In each single run, we process $[1,max_{c}]$ chunks in parallel, generating $[1,max_{c}]$ vectors of embeddings, each with dimensionality $4 \times dim$. We iteratively process chunks from $d$ until an embedding vector is generated for each chunk and thus preserving the entire text content of $d$. 

Then, we concatenate the embedding vectors maintaining the order of the respective chunks, generating a tensor of dimensionality $(n, 4\times dim)$. We process this tensor with the RNN sequentially, capturing the dependencies between them and generating a contextually enhanced representation for each chunk.

Splitting text by token count can disrupt its flow, so we use token overlap between chunks during both training and inference to maintain continuity. This technique, similar to that used by \citet{hoang-etal-2023-viettel} but applied more broadly, helps preserve the text's natural structure.

Our token overlap algorithm can be formalized as follows. Consider the judicial decision \( d \) as the tokenized sequence \( S = \{t_1, \dots, t_k\} \), where \( k \) is the number of tokens in \( d \). We define the overlap size, \( z \), as the number of tokens each chunk shares with its adjacent neighbors. Thus, any chunk shares $\left \lfloor \frac{z}{2}\ \right\rfloor$ tokens with the previous chunk and $\left \lfloor \frac{z}{2}\ \right\rfloor$ with the subsequent one. The first and last chunks, having only one adjacent chunk, share $\left \lfloor \frac{z}{2}\ \right\rfloor$ tokens with their respective neighbors. 

\begin{figure}[h]
    \centering
    \includegraphics[width=0.5\textwidth]{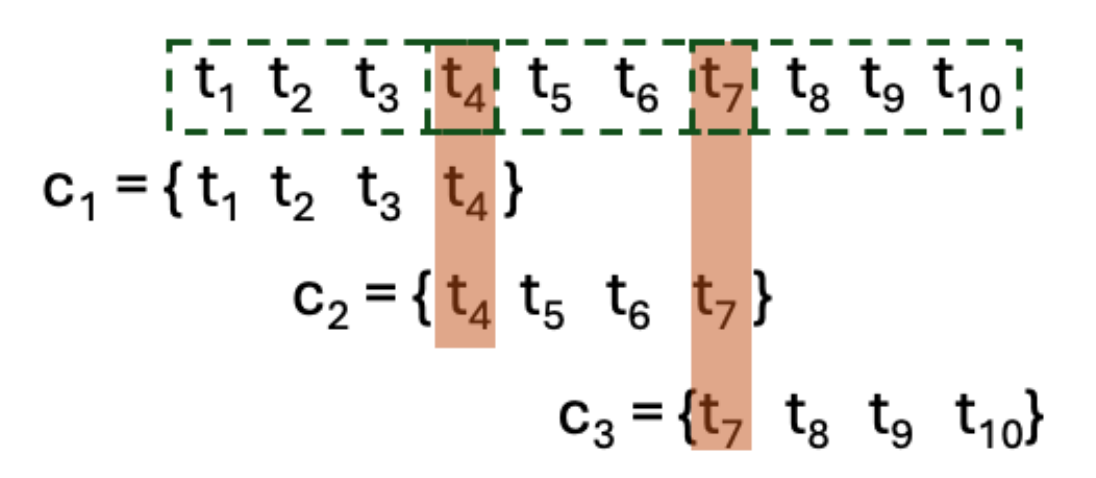}
    \caption{Overlapping chunks example.}
    \label{fig:overlap}
\end{figure}

Figure \ref{fig:overlap} provides a simple example for clarification. In this example, the chunk size is 4, and \( z = 2 \). As shown, chunk \( c_2 = \{t_4, t_5, t_6, t_7\} \) shares token \( t_4 \) with chunk \( c_1 \) and token \( t_7 \) with chunk \( c_3 \). Note that the first and last chunks share only one token with the neighboring chunk.

\section{Experiments}
\label{sec:experiments}

In this section, we design experiments to assess our proposed model, \texttt{uBERT}, and validate its effectiveness on the legal domain. We split our experiments into 3, one for each of the following research questions:

\noindent\textbf{RQ1: Would an encoder-based model benefit from using the entire text in terms of performance improvement?}

\noindent We examine the impact of processing the whole documents using multiple encoder passes. We first tested if simply increasing text chunks to process the whole text without using overlap (\texttt{uBERT\_0}) improves performance over BERT+LSTM, which processes only partial text in a single pass. Then, we investigated the effect of introducing overlaps (0 to 510 tokens\footnote{ The typical input size for BERT models is 512 tokens. Our overlap algorithm first slices the text and distributes the tokens. Only after this process are the special tokens [CLS] and [SEP] added, resulting in the well-known 512-token limit.}) between chunks to observe if the added local context enhances predictions.

\noindent\textbf{RQ2: If performance improves, does it come with reasonable computational overhead?}

\noindent We compare the inference time of our architecture agatextsuperscript all baseline models to determine if it offers a performance gain and to assess the associated computational overhead.


\noindent\textbf{RQ3: Is our architecture better for processing longer texts?}

\noindent We explore the relationship between document length and model performance. We tested the models on the full test set as well as on its subsets, the 10\% and the 1\% longest texts. This experiment involved statistical analysis to determine whether longer texts lead to better or worse predictions.



\paragraph{Data}
We used the BrCAD-5 dataset\footnote{This dataset is divided in training, validation, and test sets: the training set includes 380,673 documents, while the validation and test sets contain 76,342 and 76,299 entries, respectively.} in our experiments. 
The task is a binary classification, with Class 1 indicating that the AP reverses the previous decision, and Class 0 indicating it affirms. The dataset is imbalanced, with 22\% of the data points belonging to Class 1. Although this imbalance ratio is consistent across all dataset splits, it varies significantly with text length.

\paragraph{Models}
In this work, our model (\texttt{uBERT}) uses BERT as the encoder and LSTM as the RNN, with $max_{tok}$ set to 512 tokens and up to 15 chunks ($max_{c}$) processed in parallel.
Our training procedure follows the approach from \citet{brcad-5}, where we fine-tune the last layer of BERT and the LSTM. The fine-tuning is conducted for 1 epoch utilizing the One Cycle learning rate scheduler. Our inference procedure mirrors the training process. 

\paragraph{Baselines} our baseline models are ULMFiT (forward, backward and bidirectional)\footnote{ ULMFiT incorporates a forward language model (predicting the next token), a backward language model (predicting the previous token), and a bidirectional model that combines the two, allowing it to capture contextual information in both directions.}, Big Bird and BERT+LSTM. Notably, only ULMFiT and \texttt{uBERT} process the full text.

\paragraph{Computational Infrastructure and Resources} the experiments were conducted using an NVIDIA A100 GPU with 40 GB of RAM. Times are reported according to this hardware.

\paragraph{Evaluation Metrics}
We evaluate all models using the Macro F1 score and Matthews Correlation Coefficient (MCC). The Macro F1 score is a well-established metric across NLP fields, representing the harmonic mean of precision and recall, while MCC, though less common, is frequently used in the Legal Judgment Prediction (LJP) subfield, as noted by \citet{LJP-survey}. MCC measures the correlation between predicted and actual classifications by accounting for true positives, true negatives, false positives, and false negatives, making it suitable for imbalanced classes\footnote{ MCC ranges from -1 to 1, where 1 indicates perfect prediction, 0 indicates no better than random chance, and -1 indicates total disagreement between prediction and observation.}. Additionally, MCC is the metric used by \citet{brcad-5}, making it necessary for us to use it as well for model comparison. To compare different baselines and configurations of our \texttt{uBERT} model, we employed bootstrap resampling to obtain 95\% confidence intervals, followed by Wilcoxon-Holm post-hoc analysis to assess statistical significance with $\alpha = 5\%$, following similar approaches \cite{demvsar2006statistical,nlpstattest,ferraz2021debacer}.

\section{Results}

\begin{table}[htbp]
    \centering
    \scriptsize 
    \begin{tabular}{lcccccccc}
        \toprule
        \textbf{Dataset:} & \multicolumn{2}{c}{\shortstack{\textbf{Full Test Set} \\\textit{(76.299 documents)} \\ \textit{Imbalance Ratio = 0.28}}}
 & \multicolumn{3}{c}{\shortstack{\textbf{10\% Set} \\\textit{(7.632 documents)} \\ \textit{Imbalance Ratio = 0.32}}} & \multicolumn{3}{c}{\shortstack{\textbf{1\% Set} \\\textit{(763 documents)} \\ \textit{Imbalance Ratio = 0.54}}} \\ 
        \midrule
        \textbf{Architecture} & \textbf{Macro-F1$\uparrow$} & \textbf{MCC$\uparrow$} & \textbf{Macro-F1$\uparrow$} & \textbf{MCC$\uparrow$} & \textbf{Inf.Time$\downarrow$} & \textbf{Macro-F1$\uparrow$} & \textbf{MCC$\uparrow$} & \textbf{Inf.Time$\downarrow$}\\
        \midrule
        \textit{Baselines} \hfill \\
        ULMFiT - fwd & 65.1 \% & 0.32 & 64.9 \% & 0.32 &1h 18min & 72.3 \% & 0.47 & 11min:22s \\
        ULMFiT - bwd & 65.7 \% & 0.35 & 63.4 \% & 0.35 & 1h 18min& 59.9 \% & 0.33 &14min:44s \\
        ULMFiT - bidir & 66.9 \% & 0.37 & 64.8 \% & 0.34 & 1h 18min& 69.3 \% & 0.43 &14min:44s \\
        Big Bird & 52.0 \% & 0.27 & 44.0 \% & 0.23 & 22min & 30.0 \% & 0.08 & 2min:58s\\
        BERT+LSTM & 64.1 \% & 0.33 & 63.2 \% & 0.31 & 12min& 64.0 \% & 0.36 & 1min:29s\\ \hline
        \textit{Ours} \hfill \\
        uBERT\_0 & 63.9 \% & 0.33 & 62.6 \% & 0.31 &13min & 61.3 \%  & 0.34 & 1min:51s\\
        uBERT\_150 & 63.3 \% & 0.32 & 62.2 \% & 0.30 & 14min& 59.4 \%  & 0.32 & 2min:04s\\
        uBERT\_205 & 64.7 \% & 0.35 & 62.6 \% & 0.31 &15min & 62.0 \%  & 0.35 &2min:09s \\
        uBERT\_300 & 64.7 \% & 0.35 & 63.0 \% & 0.31 & 17min& 63.0 \%  & 0.36 & 2min:23s\\
        uBERT\_408 & 64.0 \% & 0.33 & 63.2 \% & 0.32 & 19min& 64.3 \%  & 0.38 & 2min:42s\\
        uBERT\_510 & 64.6 \% & 0.35 & 63.0 \% & 0.31 &21min & 64.2 \%  & 0.38 & 3min:08s\\
        \bottomrule
    \end{tabular}
    \vspace{5pt}
    \caption{uBERT performance across various overlap sizes compared with baselines.}
    \label{tab:overlap_range}
\end{table}

\begin{figure}  
    \includegraphics[width=\linewidth]{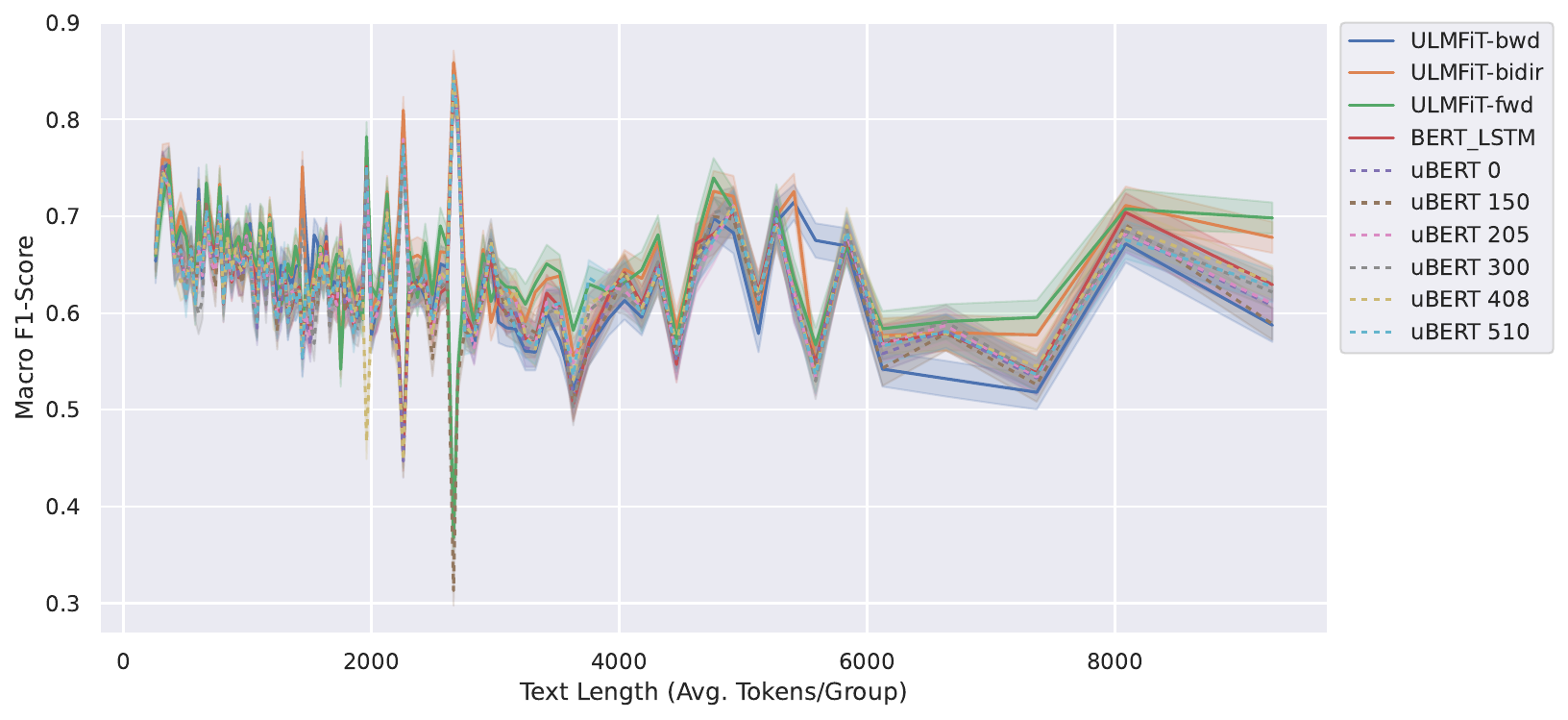}
\caption{Macro-F1 score x Avg. Tokens/Group across different groups of same size ranked by the length. The error bars represent 95\% confidence intervals obtained with bootstrap resampling.}
\label{fig:f1_length}
\end{figure}


\begin{figure}

        \begin{subfigure}[b]{0.33\textwidth}
                \includegraphics[width=\linewidth]{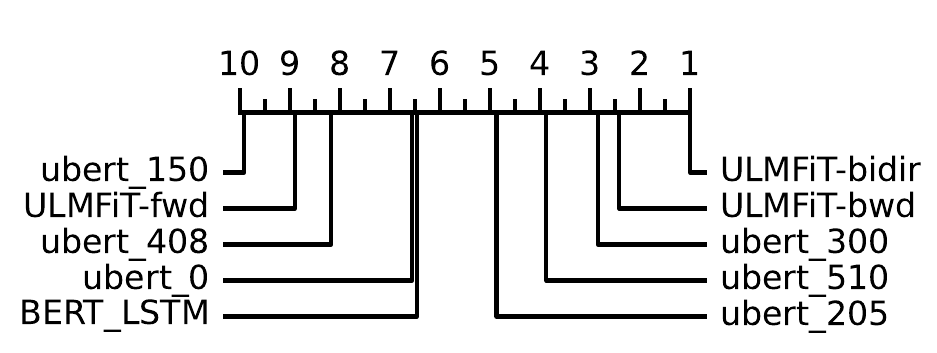}
                \captionsetup{font = scriptsize}
                \caption{\textbf{All Text}}
        \end{subfigure}%
        \hfill
        \begin{subfigure}[b]{0.33\textwidth}
                \includegraphics[width=\linewidth]{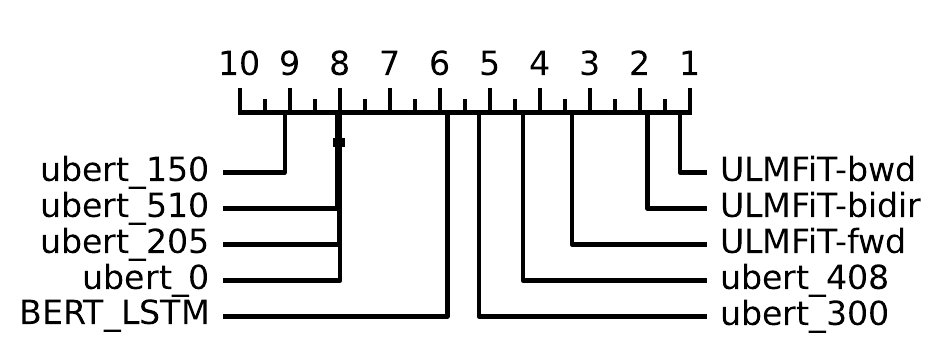}
                \captionsetup{font = scriptsize}
                \caption{\textbf{10\% Longest Text}}
        \end{subfigure}%
        \hfill
        \begin{subfigure}[b]{0.33\textwidth}
                \includegraphics[width=\linewidth]{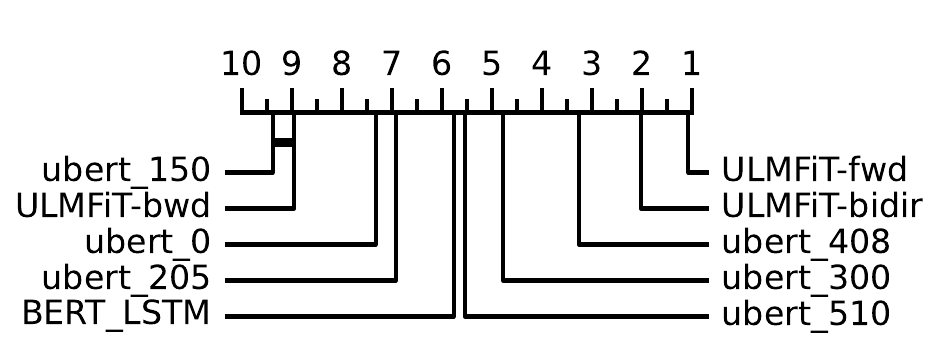}
                \captionsetup{font = scriptsize}
                \caption{\textbf{1\% Longest Text}}
        \end{subfigure}%
        \caption{Critical difference diagram showing pairwise statistical comparison between baselines and varying overlap sizes for uBERT using the MCC. Connecting bars represent no statistical difference between methods. }
        \label{fig:wilcoxon}
\end{figure}


Table \ref{tab:overlap_range} presents the results for all model configurations on the full test dataset, as well as the 10\% and 1\% longest texts. The baseline models were not run on the full test set in this study due to computational resource limitations. The results reported here are reproduced from \citet{brcad-5}, which is why Table \ref{tab:overlap_range} does not include inference times for the full test set. Figure \ref{fig:f1_length} displays the macro F1-scores across varying text lengths, while Figure \ref{fig:wilcoxon} ranks the models using the MCC metric. Although MCC is effective for within-dataset comparisons, it is less suitable across datasets with differing class imbalance; hence, we rely on the macro F1-score for cross-dataset comparisons.

\paragraph{Processing the Full Text Requires Overlap} $\enspace$ Comparing the BERT+LSTM baseline, which middle-truncates text when it exceeds input size, with our \texttt{uBERT} without overlap (\texttt{uBERT\_0}), which uses the full text, we found that \texttt{uBERT} either underperformed or matched the baseline across all metrics. Notably, it performed worse on the 1\% longest texts, where middle-truncation by BERT+LSTM occurs. \textbf{This suggests that merely processing the entire text is insufficient for longer inputs}. We hypothesize that non-overlapping chunks introduce noise due to abrupt segmentation, which degrades performance. Our results support this, showing that introducing overlap in \texttt{uBERT} configurations improves both Macro-F1 and MCC scores. The following \texttt{uBERT} configurations outperformed BERT+LSTM with statistical significance: \texttt{uBERT\_300}, \texttt{uBERT\_510} and \texttt{uBERT\_205} on full test set; \texttt{uBERT\_408} and \texttt{uBERT\_300} on 10\% longest; and \texttt{uBERT\_408}, \texttt{uBERT\_300} and \texttt{uBERT\_510} on 1\% longest. Thus, \textbf{incorporating overlap is crucial for maintaining semantic consistency and improving performance on longer texts}.

\paragraph{uBERT with Overlap is still Significantly Faster than ULMFiT} $\enspace$ As expected, introducing overlap in the \texttt{uBERT} architecture increased computational time overhead. However, across the full dataset and the 10\% longest texts, \texttt{uBERT} configurations delivered better results than the BERT+LSTM baseline with comparable inference times. Notably, uBERT\_408 achieved a 4x faster inference than ULMFiT on the 10\% longest texts. For the 1\% longest texts, the increased length required two passes of \texttt{uBERT\_408}\footnote{ With zero overlap, uBERT can process a maximum of 7,650 tokens in a single encoder pass. This limit arises because uBERT handles up to 15 chunks of 510 tokens each (excluding special tokens [CLS] and [SEP]). Therefore, documents longer than 7,650 tokens require at least two encoder passes.}, resulting in 1.8x slower inference compared to BERT+LSTM, which needed to middle-truncate in all cases. Despite this, \texttt{uBERT\_408} slightly outperformed BERT+LSTM, narrowing the performance gap with ULMFiT while maintaining a faster inference, highlighting the efficiency and effectiveness of our approach. In summary, in all subsets, \textbf{uBERT configurations reduced the BERT+LSTM gap being significantly faster than ULMFiT}. 

\paragraph{ULMFiT Outperforms uBERT on Longer Texts} $\enspace$ As shown in Figure \ref{fig:f1_length}, model differences become more clear with increasing text length. Big Bird consistently underperforms on longer texts, which is why it was excluded from the comparison charts. \textbf{While some uBERT configurations outperform BERT+LSTM on longer texts, F1 scores in both models degrade compared to full test dataset performance}. In contrast, ULMFiT models improve on longer texts compared to the full dataset. This suggests that \textbf{our architecture mitigates the degradation for longer text that is inherent to the BERT+LSTM approach, but still falls short of ULMFiT models, that handle better longer text but at a cost of 4x slower inference time}.

\section{Conclusion and Future Work}

Our experiments demonstrate that the \texttt{uBERT} model improves the handling of legal texts compared to baseline encoder-based models, particularly on longer texts, due to its capability to process entire documents using overlapping chunks. Despite the increased computational overhead, \texttt{uBERT} remains faster than ULMFiT. \texttt{uBERT} slightly outperforms BERT+LSTM, but still falls short of ULMFiT. Thus, further refinement is needed to fully match ULMFiT`s performance.
Notably, even ULMFiT, the top-performing model in our experiments, achieves relatively low Macro-F1 scores, suggesting that processing the full text alone is insufficient for high performance on this task. In this direction, future research should expand the evaluation methodology by analyzing correctly and incorrectly classified cases across all tested models to assess whether specific characteristics of judicial decisions make them more prone to misclassification by certain models. Such an analysis, however, requires a multidisciplinary approach, including expert input from highly skilled legal practitioners. 

Future research should also explore different chunking strategies to enhance text processing. Comparing syntactic chunking, which is based on grammatical structure, with semantic chunking, which is based on content meaning, could provide valuable insights. As this study focuses on a Portuguese-language dataset, evaluating these chunking approaches across datasets in multiple languages would help determine if optimal chunking strategies vary with language, contributing to more robust long-text segmentation and model performance across diverse linguistic contexts.

\section*{Acknowledgements}
This work is partially supported by the Brazilian National Council for Scientific and Technological Development (CNPq) under grant numbers 310085/2020-9, and the Brazilian Coordination for the Improvement of Higher Education Personnel (CAPES, Finance Code 001). Israel Campos Fama is funded by the Secretaria de Fazenda do Estado do Rio Grande do Sul (Sefaz-RS). Bárbara Dias Bueno is funded by the \textit{Programa Unificado de Bolsas (PUB)} undergraduate research scholarship from Universidade de São Paulo, under project 2117/2023 \textit{"Machine Learning for Text Classification in Low-Resource Scenarios"}. 

\bibliographystyle{apalike}
\bibliography{sbc-template}

\end{document}